\documentclass{article}

\usepackage{graphicx}
\usepackage{amssymb}
\usepackage{epstopdf}

\usepackage{times,graphics,amsmath,amssymb,lipsum}
\usepackage[utf8]{inputenc}
\usepackage[T1]{fontenc}
\usepackage[left=1cm, right=2cm, top=1.5cm, bottom=1.2cm]{geometry}
\providecommand {\norm} [1] {\lVert #1 \rVert}

\title{Hierarchical Graphical Models for Multigroup Shape Analysis using Expectation Maximization with Sampling in Kendall's Shape Space}

\author{	Yen-Yun Yu, P. Thomas Fletcher and Suyash P. Awate
		\\
		\\Scientific Computing and Imaging (SCI) Institute
		\\
		\\University of Utah.}
\date{Dec. 15. 2012}                                           % Activate to display a given date or no date

\begin{document}
\maketitle
%\section{}
%\subsection{}

\begin{abstract}
  
  % hierarchical model, shape
  This paper proposes a novel framework for multigroup shape analysis relying on a \emph
  {hierarchical} graphical statistical model on shapes within a population. The framework represents
  individual \emph {shapes} as pointsets modulo translation, rotation, and scale, following the
  notion in Kendall's shape space.  While individual shapes are derived from their group shape
  model, each group shape model is derived from a single population shape model. The hierarchical
  model follows the natural organization of population data and the top level in the hierarchy
  provides a common frame of reference for multigroup shape analysis, e.g. classification and
  hypothesis testing.
  % generative model
  Unlike typical shape-modeling approaches, the proposed model is a \emph {generative} model that
  defines a joint distribution of object-boundary data and the shape-model variables.
  % correspondence problem
  Furthermore, it naturally enforces optimal correspondences during the process of model fitting and
  thereby subsumes the so-called \emph {correspondence problem}.
  % EM
  The proposed inference scheme employs an expectation maximization (EM) algorithm that treats the
  individual and group shape variables as hidden random variables and integrates them out before
  estimating the parameters (population mean and variance and the group variances).
  % sampling
  The underpinning of the EM algorithm is the \emph {sampling of pointsets, in Kendall's shape
    space}, from their posterior distribution, for which we exploit a highly-efficient scheme based
  on Hamiltonian Monte Carlo simulation.
  % hypothesis testing
  Experiments in this paper use the fitted hierarchical model to perform (1) \emph{hypothesis
    testing} for comparison between pairs of groups using permutation testing and (2)
  \emph{classification} for image retrieval.
  % results
  The paper validates the proposed framework on simulated data and demonstrates results on real
  data.
  
\end{abstract}

%%%%%%%%%%%%%%%%%%%%%%%%%%%%%%%%%%%%%%%%%%%%%%%%%%%%%%%%%%%%%%%%%%%%%%%%%%%%%%% 
%% INTRODUCTION
%%%%%%%%%%%%%%%%%%%%%%%%%%%%%%%%%%%%%%%%%%%%%%%%%%%%%%%%%%%%%%%%%%%%%%%%%%%%%%% 

\section{Introduction and Related Work}

%%% shape analysis, applications
Shape modeling and analysis is an important problem in a variety of fields including biology, medical image analysis, and computer
vision\cite{Bookstein1978,MardiaBook,Goodall1991,Kendall1984} that has received considerable attention over the last few decades.
Kendall\cite{Kendall1984} defines \emph {shape} as an equivalence class of pointsets under the similarities generated by translation, rotation, and
scaling.
% applications
Objects in biological images or anatomical structures in medical images often possess shape as the sole identifying characteristic instead of color or
texture.  Applications of shape analysis beyond natural contexts include handwriting analysis and character recognition. In the medical context,
shapes of human anatomical structures can provide crucial cues in diagnosis of pathologies or disorders. The key problems in this context lie in the
fitting of shape models to population image data followed by statistical analysis such as hypothesis testing to compare groups, classification of
unseen shapes in one of the groups for which the shape models are learnt, or unsupervised clustering of shapes.

%%% hierarchical models of shape: population, groups, individuals
A variety of applications of shape analysis deal with population data that naturally calls for a \emph {hierarchical} modeling approach.
% natural organization
First, the hierarchical model follows the natural organization or structure of the data. For instance, as in Figure~\ref{fig:model}, the model
variables at the top level capture statistical properties of the population (e.g., individuals with and without medical conditions), the model
variables at a lower level capture statistical properties of different groups within a population (e.g., clinical cohorts based on gender or age or
type of disease within a spectrum disorder), and the model variables at the lowest level capture individual properties, which finally relate to the
observed data.
% hypothesis testing, classification
Second, the top-level population variables are necessary to enable comparison (through hypothesis testing or classification) between the groups within
the population because the population shape variable helps provide a common reference frame (e.g., a grand mean) for the shape models underlying
different groups.

%%% previous hierarchical models
Previous works on ``hierarchical'' (overloaded term) shape modeling concern (i)~multi-resolution/scale models e.g., face model at fine-to-coarse
resolutions) or (ii)~multi-part models (e.g., car decomposed into body, tires, hood, trunk) with inter-part spatial relationships. In contrast, our
proposed model is the first hierarchical probabilistic model comprising a population with multiple groups (e.g., vehicle population comprising sedans,
trucks, buses).

%%% shape model, pointset
In the context of image-based applications, shape models are constructed from image-derived data that is typically in the form of a set of pixels
lying close to the object boundary detected in the set of images; each pointset has the same number of points (also interpreted as \emph
{landmarks}\cite{MardiaBook}). To obtain valid and useful shape models from the data, it is crucial to determine (i)~the positions of the points,
within each pointset, in relationship to the object boundary in the image, and (ii)~meaningful \emph {correspondences}\cite{MardiaBook} of points
across pointsets in the group and the population.
% landmarks given
Some of the early approaches to shape modeling\cite{Cootes1995,Goodall1991} assumed knowledge of a set of homologous landmarks, typically manually
defined, comprising the pointset for each object in the population. They proposed methods for learning Gaussian shape models for such landmark data
observed for a population. However, for shapes that are complex or 3 dimensional, manually defining landmarks can be infeasible and inaccurate. Thus,
there is a need to include the pointsets, describing each individual shape in the population, as variables in an optimization process.
% points given, correspondences optimized
Several approaches to the correspondence problem attempt to find the \emph {simplest} explanation for the point positions, including methods based on
the logarithm of the determinant of the model covariance matrix\cite{Kotcheff1998}, minimum description length\cite{Davies2002,Thodberg2003}, and
minimum entropy\cite{Cates2008a,Oguz2009}. While these model-building criteria are based on statistical properties of the pointsets, (i)~none of
these approaches use \emph {generative} statistical models, (ii)~they force correspondences in an adhoc manner by introducing adhoc terms in the
objective function, and (iii)~have 2 independent stages where the first finds optimal correspondences and the second estimates model parameters.

%%% previous shape models
Some generative model approaches to shape have been introduced in the computer vision
literature\cite{Coughlan2002,Gu2008,Rangarajan2003,Neumann2003,Zhu1999}, but these are defined relative to some pre-determined template shape with
manually-placed landmarks.
% our approach
On the other hand, the proposed method learns shape models from populations of detected object boundaries in images, without any a priori definition
of a template or known landmarks. Furthermore, it proposes a generative statistical model for hierarchical shape modeling that naturally enforces
optimal correspondences as a result of model fitting. In this way, the proposed modeling and optimization framework subsumes the correspondence
problem.

%%% contributions

This paper makes several contributions.
% model: hierarchical generative model
First, it proposes a novel hierarchical generative statistical model for modeling shapes within a population. This model tightly couples each
individual shape model to the observed data (i.e., the set of points on the object boundary detected in the image) by designing their joint
distribution using the (nonlinear) \emph {current norm} between pointsets\cite{Joshi2011,Vaillant2005}.
% optimization: EM
Second, the proposed optimization framework employs an \emph {expectation maximization} (EM) algorithm\cite{Dempster77} that treats the individual
and group shape variables as hidden random variables and integrates them out before estimating the parameters, which are population mean and variance
and the group variances. In this way, the proposed Bayesian EM-based inference framework is an improvement over using the mode approximation for the
hidden variables.
% optimization: sampling shapes using HMC
Third, the underpinning of the EM algorithm is the sampling of shapes (within Kendall shape space) from their posterior PDFs, e.g., shape variables
for individual data have posterior PDFs involving (i)~likelihood PDFs designed using the (nonlinear) current norm and (ii)~prior PDFs conditioned on
the group shape mean. For this purpose, this paper exploits a novel highly-efficient sampling scheme relying on \emph {Hamiltonian Monte Carlo} (HMC)
simulation\cite{Duane1987,Neal1994}\cite{NealBook}\cite{Neal2010}.

%%% organization

The remainder of the paper is organized as follows.
Section~\ref{sec:model} gives an overview of the proposed hierarchical statistical generative model for multigroup population population data.
Section~\ref{sec:opt} presents the EM formulation for the optimization problem. It defines the hidden variables and the parameters and the joint PDF,
the details of the expectation step for integrating over the hidden variables, involving the HMC approach, and the subsequent maximization step to
optimize the parameters.
Section~\ref{sec:testing} describes a hypothesis-testing scheme, using permutation testing, for comparing any pairs of groups within the population.
Section~\ref{sec:classification} presents a method for shape classification.
Section~\ref{sec:conclusion} summarizes the paper.

\section {Hierarchical Multigroup Shape Model}
\label {sec:model}

\begin{figure*}[!t]
  \begin{center}
    \centerline{\includegraphics[scale=0.25]{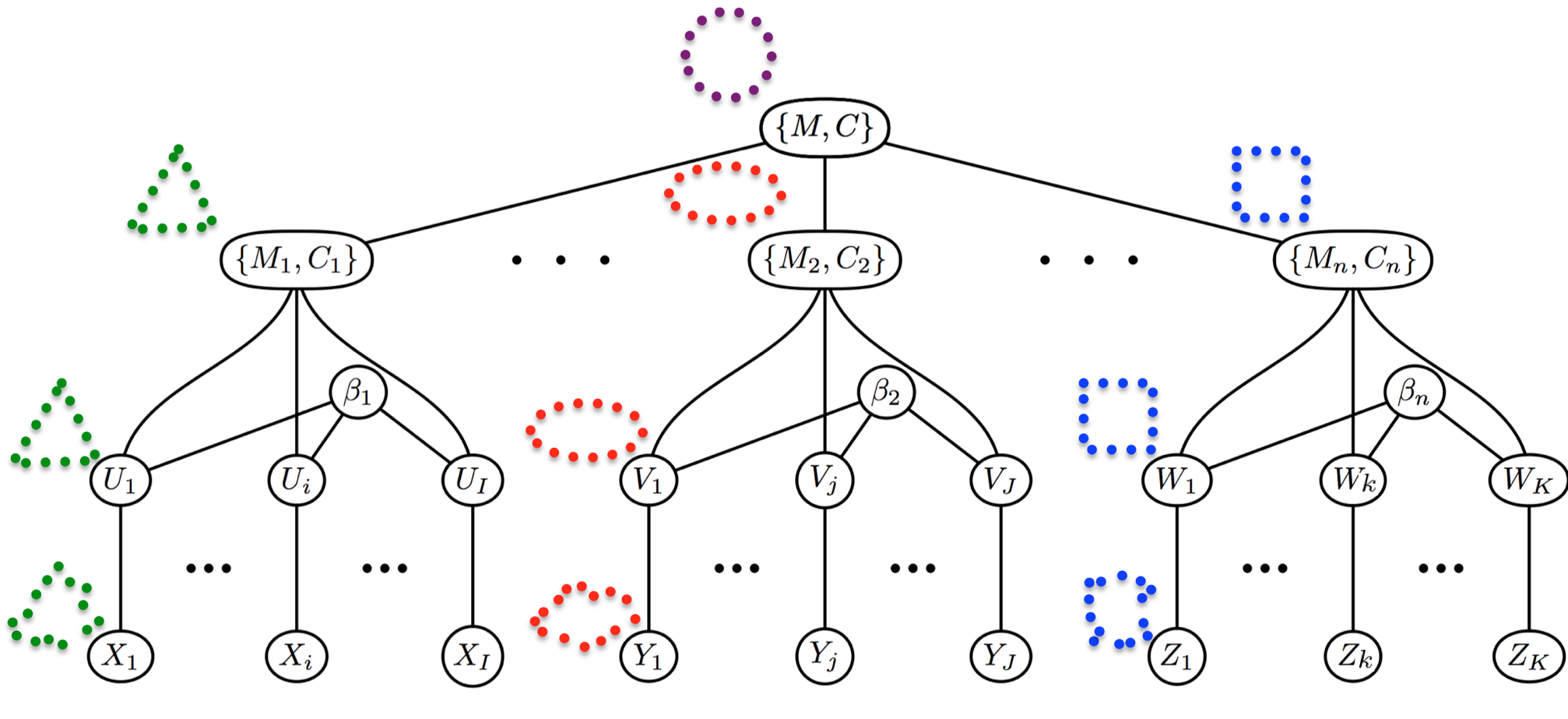}}
    \vskip -0.1in
    \caption
    {
      {\bf Hierarchical Modeling of Multiple Groups of Shapes in a Population.} 
    }
    \label {fig:model}
  \end{center}
  \vskip -0.2in
\end{figure*}

This section gives an overview of the proposed hierarchical statistical generative model for multigroup population population data, illustrated in
Figure~\ref{fig:model}.

% model: one group
Let the group of random variables $X \triangleq \{ X_i \}_{i = 1,\ldots,I}$ model \emph{observed
  data} where each (vector) random variable $X_i$ is a set of observed (corrupted) points
(representing shape) on the boundary of the $i^{\text{th}}$ structure. That is, $X_i \triangleq \{
X_i(t) \}_{t=1,\ldots,T}$ where $X_i(t) \in \mathbb{R}^D$ is the $D$-dimensional coordinate of the
$t^{\text{th}}$ point on the boundary of the $i^{\text{th}}$ structure. We assume that all $I$
random variables $X_i$ belong to a \emph {single group} of shapes, with the group's shape model
given by variables $\{ M_1, C_1 \}$ where $M_1$ is the mean shape and $C_1$ the covariance of the
shapes in the group.

% model: multiple groups
Similarly, we have other groups of shapes modeled by analogous random variables, e.g., group $Y
\triangleq \{ Y_j \}_{j = 1,\ldots,J}$ with group mean and covariance $\{ M_2, C_2 \}$, group $\{
Z_k \}_{k = 1,\ldots,K}$ with group mean and covariance $\{ M_3, C_3 \}$, etc.
In order to be able to perform shape classification or hypothesis testing between groups, we ensure
that the shape models lie in the same space by enforcing the same number of points $T$ in all group
shape means, i.e., $M_1 \triangleq \{ M_1(t) \}_{t=1,\ldots,T}$, $M_2 \triangleq \{ M_2(t)
\}_{t=1,\ldots,T}$, etc. all have the same number of points $T$. Thus, $M_1 \in \mathbb{R}^{TD}$,
$M_2 \in \mathbb{R}^{TD}$, $M_3 \in \mathbb{R}^{TD}$, $\forall i \text { } X_i \in \mathbb{R}^{TD}$,
$\forall j \text { } Y_j \in \mathbb{R}^{TD}$, and $\forall k \text { } Z_k \in \mathbb{R}^{TD}$.
% model: population
Furthermore, we assume that all groups belong to a single population of shapes with shape parameters
$\{ M, C \}$ that represent the mean and covariance on the group means $M_1,M_2$, etc.

% problem
In the rest of the formulation, without loss of generality, we consider two groups for simplicity.
For each group of shapes, we need to model the joint PDF comprising the population model variables $M,C$, group-level model variables $M_1,C_1,
M_2,C_2$, and the data $X,Y$, i.e., $P (M,C, M_1,C_1, M_2,C_2, X, Y)$.
Given observed data $x,y$ that comprises sets of shapes within 2 groups, the problem addressed in this paper is to fit the aforementioned hierarchical
multigroup shape model to the data and subsequently perform hypothesis testing or classification.

\section{Inference via Monte-Carlo Expectation Maximization}
\label {sec:opt}

We solve the multigroup shape-fitting problem using the EM framework\cite{Dempster77}. Figure~\ref{fig:model} gives an overview of the approach.

% hidden variables: individual level
For the first group, i.e., corresponding to $X$, we assume a \emph{hidden} random variable $U
\triangleq \{ U_i \}_{i = 1,\ldots,I}$ where each (vector) random variable $U_i$ represents the
Kendall shape corresponding to the observed data $x_i$. The number of points in $U_i$ is the same as
those in $M_1$, i.e., $T$ points. Similarly, we assume a hidden variable $V$ for the second group.
% hidden variables: group level
We also consider the group means $M_1, M_2$ as hidden (Kendall-shape) variables.
% parameters
The \emph{parameters} in our model include the group-specific covariances $C_1, C_2$, the population
mean $M$ and the population covariance $C$. We introduce parameters $\beta_1,\beta_2$, one per
group, to control the smoothness of the hidden shape variables $U_i,V_j$. In this paper,
(i)~$\beta_1,\beta_2$ are fixed to 1 and (ii)~smoothness parameters are ignored for the group means
because of their limited practical utility. Let $\theta \triangleq \{ C_1, C_2, M, C \}$ be the set
of parameters to be estimated.
% 2 groups
Then, the inference problem is:
\begin{align}
  &
  \max_{\theta}
  P (x,y | \theta)
  =
  \nonumber \\
  &
  \max_{\theta}
  \int
  P (U,V, M_1,M_2, x,y | \theta) du dv dm_1 dm_2
\end{align}
In the $i^{\text{th}}$ iteration of EM optimization, with the parameter estimate $\hat \theta^i$,
the E step constructs the \emph {Q function}
\begin{align}
  &
  Q (\theta | \hat \theta^i)
  \triangleq
  \nonumber \\
  &
  %E_{ P (U,V, M_1,M_2, x,y | \hat \theta^i ) }
 E_{ P (U,V, M_1,M_2 |  x , y , \hat \theta^i ) }
  \log
  P ( U,V, M_1, M_2, x,y | \theta ).
\end{align}
which, because of the intractability of the expectation, we approximate using Monte-Carlo simulation
as follows:
\begin{align}
  &
  Q (\theta | \hat \theta^i)
  \approx
  \hat Q (\theta | \hat \theta^i)
  \triangleq
  \sum_{s = 1}^S
  \frac {1} {S}
  \log
  P ( u^s,v^s, m_1^s,m_2^s, x,y | \theta ),
  \nonumber \\
  &
  \text { where }
  (u^s,v^s, m_1^s,m_2^s)
  \sim
  P (U,V, M_1,M_2 | x,y, \hat \theta^i)
\end{align}
and where shapes $u,v$, which correspond to each observed sets of boundary points $X,Y$, as well as
shapes $m_1,m_2$, which correspond to the group means, are sampled from the posterior $P (U,V,
M_1,M_2 | x,y, \hat \theta^i)$.

% model
In the EM framework, with the hidden variables $U,V$, we model the joint PDF comprising the population shape variables, per-group shape variables,
per-individual shape variables, and the data as:
\begin{align}
  &
  P (M,C, M_1,C_1, M_2,C_2, U,V, X, Y)
  \triangleq
  \nonumber \\
  &
  P (M,C,C_1,C_2)
  P (M_1 | M,C)
  P (M_2 | M,C)
  \nonumber \\
  &
  \Pi_{i = 1}^I
  P (U_i | M_1,C_1,\beta_1)
  P (X_i | U_i)
  \nonumber \\
  &
  \Pi_{j = 1}^J
  P (V_j | M_2,C_2,\beta_2)
  P (Y_j | V_j), \text { where}
\end{align}
\vskip -0.2in
\begin {itemize}
  \vskip -0.4in
\item $P (M,C,C_1,C_2)$ is a prior on the population shape mean and covariance and the per-group shape covariances (ignored in this paper).
  \vskip -0.4in
\item $P (M_1 | M,C), P (M_2 | M,C)$ are modelled as Gaussians with mean $M$ and covariance $C$.
  \vskip -0.4in
\item We design $P (U_i | M_1,C_1,\beta_1)$ (and similarly $P (V_j | M_2,C_2,\beta_2)$) so as to penalize (i)~the Mahalanobis distance of $U_i$ under
  $M_1,C_1$ as well as the (ii)~deviation of every individual point $U_i(t)$ given its neighbors. For structures in 2-dimensional images ($D = 2$ in
  this paper) whose boundaries are planar closed curves, the neighborhood system underlying shape variable $U_i$ (which comprises points $\{ U_i(t)
  \}$) assigns two neighbors to each point: i.e., neighbors of $t: \forall 2 \le t \le (T-1)$ are $(t-1)$ and $(t+1)$, neighbors of $t=1$ are $t=T$
  and $t=2$, and neighbors of $t=T$ are $t=(T-1)$ and $t=1$. Given this neighborhood system $\mathcal{N} \triangleq \{ (t_1,t_2): t_1, t_2 \text { are
    neighbors } \}$,
  \begin{align}
    &
    P (U_i | M_1,C_1,\beta_1)
    \triangleq
    \nonumber \\
    \frac
    {1}
    {Z}
    \exp
    \bigg(
    &
    - 0.5
    (U_i-M_1)' C_1^{-1} (U_i-M_1)
    \nonumber \\
    &
    -
    \beta_1
    \sum_{ (t_1,t_2) \in \mathcal{N} }
    \norm { U_i(t_1) - U_i(t_2) }^2
    \bigg)
  \end{align}
  where $\beta_1$ controls the level of smoothness ($\beta_1 = \beta_2 = 1$ in this paper) and $Z$ is the partition function.
  \vskip -0.4in
\item $P (X_i | U_i), P (Y_j | V_j)$ are modelled using the current norm as follows.
  % data fit: current norm
  Consider two pointsets $A = \{ a_i \}_{i=1,\ldots,I}$ and $B = \{ b_j \}_{j=1,\ldots,J}$. Given a kernel similarity function $K(\cdot)$, which is
  Gaussian in this paper, the squared current norm between $A$ and $B$ is:
  \begin{align}
    &
    d_k^2 (A, B)
    \triangleq
    \sum_{i'=1}^I
    \sum_{i''=1}^I
    K (a_{i'}, a_{i''})
    +
    \nonumber \\
    &
    \sum_{j'=1}^J
    \sum_{j''=1}^J
    K (b_{j'}, b_{j''})
    -
    2
    \sum_{i=1}^I
    \sum_{j=1}^J
    K (a_i, b_j).
  \end{align}
  We exploit the current norm to define a PDF on data shapes $X_i$ given shape variables $U_i$ as follows:
  \begin{align}
    P (X_i | U_i)
    \triangleq
    \frac
    {1}
    {\gamma}
    \exp
    \Big(
    -
    d_k^2 (X_i, U_i)
    \Big),
  \end{align}
  where $\gamma$ is a suitable normalization factor. Note: this allows the number of points in the
  shape models to be different from the number of observed boundary points.
\end {itemize}

\subsection {E Step: Sampling in Shape Space using Hamiltonian Monte Carlo}

In EM optimization, the E step requires sampling the hidden variables $U,V, M_1,M_2$ from the posterior PDF $P (U,V, M_1,M_2 | x,y, \hat \theta^i)$.
We employ the Hamiltonian Monte Carlo procedure\cite{Duane1987}\cite{NealBook}\cite{Neal2010} for this sampling. HMC is a highly efficient sampling
technique that belongs to the class of Markov-chain Monte-Carlo samplers. HMC exploits the gradient of the energy function (i.e., $\log P (U,V,
M_1,M_2 | x,y, \hat \theta^i)$, in our case) for fast exploration of the space of the hidden random variables. The key idea underlying HMC is to first
augment the original variables with auxiliary momentum variables, then define a Hamiltonian function combining the original and auxiliary variables,
and, subsequently, alternate between simple updates for these auxiliary momentum variables and Metropolis updates for the original variables. HMC
proposes new states by computing a trajectory according Hamiltonian dynamics implemented with a leapfrog method. Such new states can be very distant
from the current states, thereby leading to a fast exploration of the state space. Furthermore, Hamiltonian dynamics theoretically guarantees that the
new proposal states are accepted with high probability.
We sample the set of hidden variables $U,V, M_1,M_2$ by successively sampling the variables $\{ U_i \}, \{ V_i \}, \{ M_1 \}, \{ M_2 \}$, following a
Gibbs sampling technique, where each variable is sampled using HMC. The HMC procedure requires gradients of the energy function $\log P (U,V, M_1,M_2
| x,y, \hat \theta^i)$ with respect to the hidden variables $\{ U_i \}, \{ V_j \}, M_1,M_2$. The gradient of $\log P (U,V, M_1,M_2 | x,y, \hat
\theta^i)$ with respect to the group mean shape variables $M_1,M_2$ as well as the shape variables $U_i,V_j$ is straightforward.

\begin{figure}[!t]
  \begin{center}
    \centerline{\includegraphics[width=0.8\columnwidth]{hyperSphere.mps}}
    \centerline{\includegraphics[width=0.6\columnwidth]{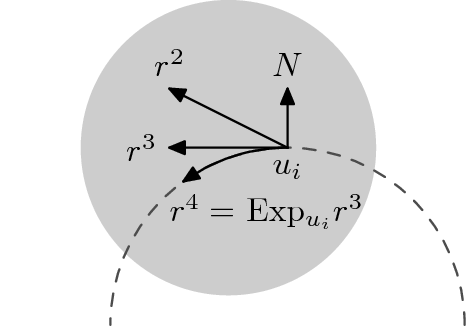}}
    \vskip -0.1in
    \caption
    {
      {\bf Sampling in Kendall Shape Space via Gradient Projection within HMC.}
      {\bf Top: } shows an illustration of Kendall's \emph {pre-shape} space\cite{Kendall1984}
      (dotted hypersphere), which is the intersection of the (bold) hypersphere of fixed radius
      $\rho$ (i.e., $\sum_t \norm{u_i(t)}^2 = \rho^2$; fixes scale) and the hyperplane through the
      origin (i.e., $\sum_t u_i(t) = 0$; fixes translation).
      % remove translation effects
      For a pointset $u_i$, log-posterior gradients $r^1$ are projected onto the hyperplane to remove translation effects from the HMC update.
      % remove scale
      {\bf Bottom: } The resulting projection $r^2$ is then projected onto the tangent space at
      $u_i$, on Kendall pre-shape space (dotted hypersphere), to remove scale changes. The resulting
      tangent-space projection $r^3$ is mapped to the pre-shape space via the manifold exponential
      map.
      % remove rotation
      The resulting pre-shape $r^4$ is rotationally aligned with the original shape $u_i$, yielding $r^5$ (not shown in figure).
      These steps project the log-posterior gradient at $u_i$, within HMC, to generate a new sample
      point (not shown in figure) in Kendall shape space.
    }
    \label{fig:shapeSampling}
  \end{center}
  \vskip -0.2in
\end{figure}

% gradients for HMC: individual shape, avoiding shifts, shrinkage, and rotation
{\bf Sampling in Kendall Shape Space using HMC:} In general, the gradient of the log posterior for each point $t$ in the shape $U_i$ can change the
translation, scale, and pose/rotation for the pointset. Specifically, the smoothness penalty on the shape variables, e.g., $u_i$, through $\beta_1$,
tends to move the $t^{\text{th}}$ point $u_i(t)$, to the mean of its neighbors, thereby shrinking the pointset. These effects are undesirable and
significantly reduce the efficiency of the shape-sampling scheme by preventing sampling restricted to Kendall shape space.
To avoid this effect, we propose a replace the gradient of the log posterior, within HMC, by a \emph {projected gradient} that restricts the updated
shape to Kendall shape space by removing effects of translation, scale, and rotation from the HMC update.  Figure~\ref{fig:shapeSampling} illustrates
this process. Without loss of generality, we assume that (i)~$u_i$ is centered, i.e., $\sum_t u_i(t) = {\bf 0}$, and thus lies on a hyperplane, (ii)~
$u_i$ has size $\rho$, i.e., $\sum_t \norm {u_i(t)}^2 = \rho^2$, and thus lies on a hypersphere and (iii)~$u_i$ has a certain pose.
Kendall \emph {pre-shape} space\cite{Kendall1984} is the hypersphere ($S^{T D - D}$ with radius $\rho$) that is the intersection of the hypersphere
$S^{T D}$ and the aforementioned hyperplane ($\in \mathbb{R}^{T D}$) that passes through the origin.
First, we decompose the gradient for $u_i$ into 3 components: (i)~a component orthogonal to the hyperplane (this causes translation of $u_i$ from its
current location), (ii)~a component within the hyperplane but orthogonal to the hypersphere (this causes changes in the size of $u_i$), and (iii)~a
component within the hyperplane and tangent to the hypersphere. Subsequently, we project the gradient vector to reduce it to the third component and
then take the manifold exponential map on the hypersphere. Lastly, to remove effects of rotation, we rotationally align the resulting pointset in pre-shape
space to the original pointset $u_i$. This gives the updated $u_i$ in Kendall shape space.

{\bf Sampling in Hierarchical Shape Model:} We generate a sample of size $S$ as follows ($S = 100$ in this paper):
\begin{enumerate}
\item Set the sample index variable $s$ to $0$.  Initialize the sampling procedure with the sample point $s=0$ denoted by $u^0 \triangleq \{ u_i^0 \},
  v^0 \triangleq \{ v_i^0 \}, m_1^0, m_2^0$.  Given sample point $s$, sample the $(s+1)^{\text{th}}$ sample point as follows.
\item $\forall i$ sample $u_i^{s+1} \sim P (U_i,v^s, m_1^s,m_2^s | x,y, \hat \theta^i)$ via projected-gradient HMC.
\item $\forall j$ sample $v_j^{s+1} \sim P (u^{s+1},V_j, m_1^s,m_2^s | x,y, \hat \theta^i)$ via projected-gradient HMC.
\item Sample $m_1^{s+1} \sim P (u^{s+1},v^{s+1}, M_1,m_2^s | x,y, \hat \theta^i)$ via projected-gradient HMC.
\item Sample $m_2^{s+1} \sim P (u^{s+1},v^{s+1}, m_1^{s+1},M_2 | x,y, \hat \theta^i)$ via projected-gradient HMC.
\item If $s+1 = S$, then stop; otherwise increment $s$ by $1$ and repeat the sampling steps.
\end{enumerate}
To ensure the generation of independent samples between Gibbs iteration $s$ and the next $s+1$, we run the HMC procedure sufficiently long.
Furthermore, as required in Gibbs sampling, we ensure independent samples between EM iteration $i$ and the next $i+1$ by \emph {burning in} and
discarding the first few sample points $s$ for use in the M step.

\subsection {M Step: Optimizing Parameters}

In the EM optimization, at the $i^{\text{th}}$ iteration, the M step entails maximization of the Q function $Q (\theta | \hat \theta^i)$ with respect
to the parameters $\theta$ and sets $\hat \theta^{i+1} \triangleq \arg \max_\theta \sum_{s=1}^S \log P ( u^s,v^s, m_1^s,m_2^s, x,y | \theta )$.
Subsequently, we get optimal values in closed form for the parameters $\hat C_1^{i+1},\hat C_2^{i+1}, \hat M^{i+1},\hat C^{i+1}$:
\begin{align}
  \hat C_1^{i+1}
  &
  =
  \frac {1} {S I}
  \sum_{s=1}^S
  \sum_{i=1}^I
  \big(
  (u_i^s - m_1^s) (u_i^s - m_1^s)'
  \big)
  \\
  \hat C_2^{i+1}
  &
  =
  \frac {1} {S J}
  \sum_{s=1}^S
  \sum_{j=1}^J
  \big(
  (v_j^s - m_2^s) (v_j^s - m_2^s)'
  \big)
  \\
  \hat M^{i+1}
  &
  =
  \frac {1} {2 S}
  \sum_{s=1}^S
  (
  m_1^s
  +
  m_2^s
  )
  \\
  \hat C^{i+1}
  &
  =
  \frac {1} {2 S}
  \sum_{s=1}^S
  \big(
  (m_1^s - M) (m_1^s - M)'
  +
  \nonumber \\
  &
  (m_2^s - M) (m_2^s - M)'
  \big).
\end{align}

\section {Validating the Hierarchical Shape Model and Inference on Simulated Shapes}
\label {sec:validation}

\begin{figure*}[!t]
  \begin{center}
    \scalebox{0.32}{\centerline{\includegraphics{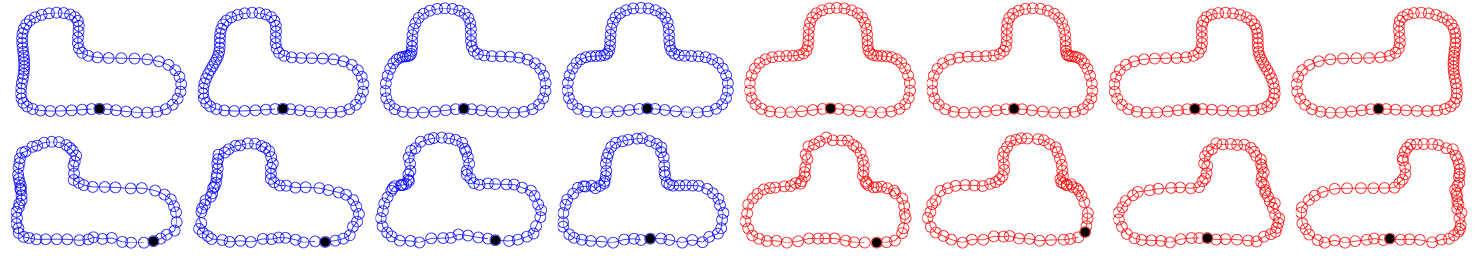}}}
    \vskip -0.1in
    \caption
    {
      {\bf Simulated Box-Bump Shapes}\cite{Kotcheff1998}.
      {\bf Top Row: } Simulated ground-truth shape models for the 2 groups that each comprise 4 shapes. The bump for the first group (blue) is on the
      left while the bump for the second group (red) is on the right. Each shape has $64$ points shown by circles. The black filled circle indicates
      the first point in the list; other points are numbered counter-clockwise.
      {\bf Bottom Row: } Corrupted data where the point ordering in each shape is randomly circularly shifted (to induce poor correspondences) and
      independent Gaussian noise is added to each point position (to mimic errors in boundary detection).
    }
    \label {fig:boxBumpData}
  \end{center}
  \vskip -0.1in
\end{figure*}

\begin{figure*}[!t]
  \begin{center}
    \scalebox{0.4}{\centerline{\includegraphics{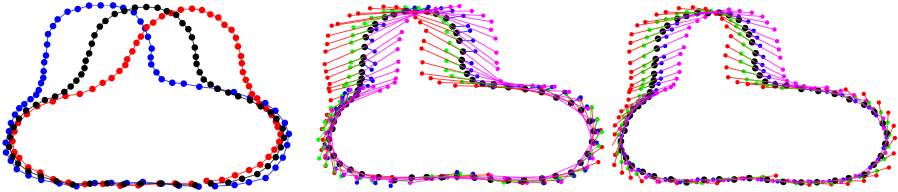}}}
    \vskip -0.1in
    \caption
    {
      {\bf Left: } shows the optimal population mean $M$ (black dots) along with the expected values for the group means $M_1$ (blue dots) and $M_2$
      (red dots) after EM inference.
      {\bf Middle: } shows the {\bf correspondence} of points for the corrupted data $x_i$ across a selected group (blue shapes).
      {\bf Right: } shows the correspondences for the expected values of shape models $U_i$ after EM inference.
    }
    \label {fig:boxBumpMeanCorres}
  \end{center}
  \vskip -0.1in
\end{figure*}

\begin{figure*}[!t]
  \begin{center}
    \scalebox{.18}{\centerline{\includegraphics{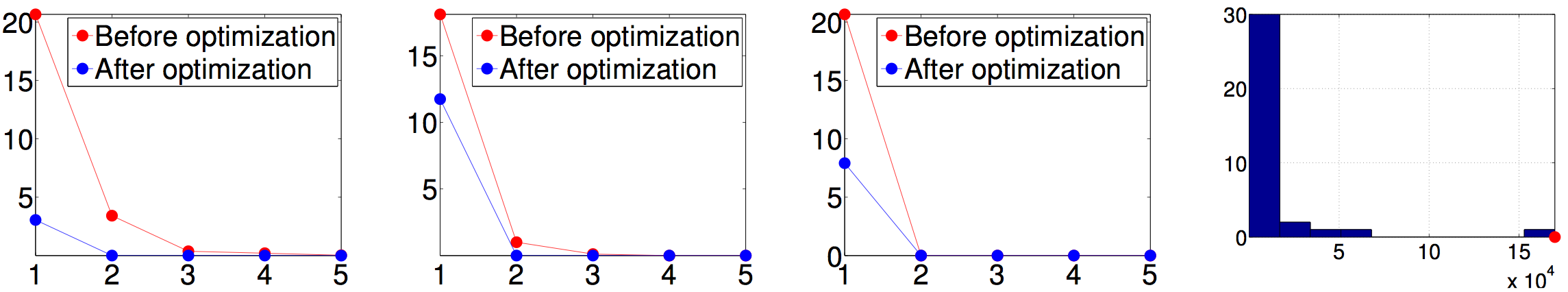}}}
    \vskip -0.1in
    \caption
    {
      {\bf 3 Graphs on Left: } show the largest $5$ eigen values of the covariance matrices $C_1$, $C$, and $C_2$, respectively, before and after EM
      optimization.
      {\bf Rightmost: } shows the histogram of Hotelling's $T^2$ statistics obtained by permutation testing (all $^8C_4 / 2 = 35$ unique group-label
      permutations) and the $T^2$ statistic for the non-permuted groups (red dot).
    }
    \label {fig:boxBumpCovTest}
  \end{center}
  \vskip -0.1in
\end{figure*} 

This section provides an example of a standard simulated test dataset\cite{Kotcheff1998} and shows that the proposed framework is able to correctly
learn the true group and population models.
% truth
Figure~\ref{fig:boxBumpData} (top row) shows simulated (ground-truth) pointsets, which are very similar to those used in\cite{Kotcheff1998}, where
the population is a collection of \emph {box-bump}\cite{Kotcheff1998} shapes where the location of the bump (and each point) exhibits linear
variation, across the group, over the top of the box. While the desired population mean $M$ corresponds to a shape with the bump exactly in the
middle, the bumps in the true group means $M_1,M_2$ are located symmetrically on either side of the middle. The true covariance matrices for the
groups ($C_1,C_2$) and the population ($C$) have a single non-zero eigen value.
% data
Figure~\ref{fig:boxBumpData} (bottom row) shows the observed corrupted data, i.e., $\{ x_i \}_{i=1}^4, \{ y_j \}_{j=1}^4$, where we induce poor
correspondences and noise in the point locations.

% result: mean
Figure~\ref{fig:boxBumpMeanCorres} shows the means for the groups and population after EM optimization. We see that the estimated population mean $M$
and the expected values of the group means $M_1,M_2$ after EM optimization are close to their true values.
% result: correspondence
Figure~\ref{fig:boxBumpMeanCorres} also shows that the correspondence of corrupted data points $x_i(t)$, across the group ($i=1,\ldots,4$), is poor,
indicating a large variance. On the other hand, after EM optimization, the correspondence of the expected values of the shape variables $U_i(t)$
indicates a significantly lower variance and correctly shows the linear variation in the point location across the group.

% result: covariance
Figure~\ref{fig:boxBumpCovTest} shows the covariances for the groups and population after EM optimization. We see that the group covariances $C_1,C_2$
and the population covariance $C$ after EM optimization indicate lower variance and a single non-zero eigen value, agreeing with the underlying true
model.

\section {Results of Hierarchical Shape Model and Inference on a Leaf Database}
\label {sec:results}

\begin{figure}[!t]
  \begin{center}
    \centerline{\includegraphics[width=\columnwidth]{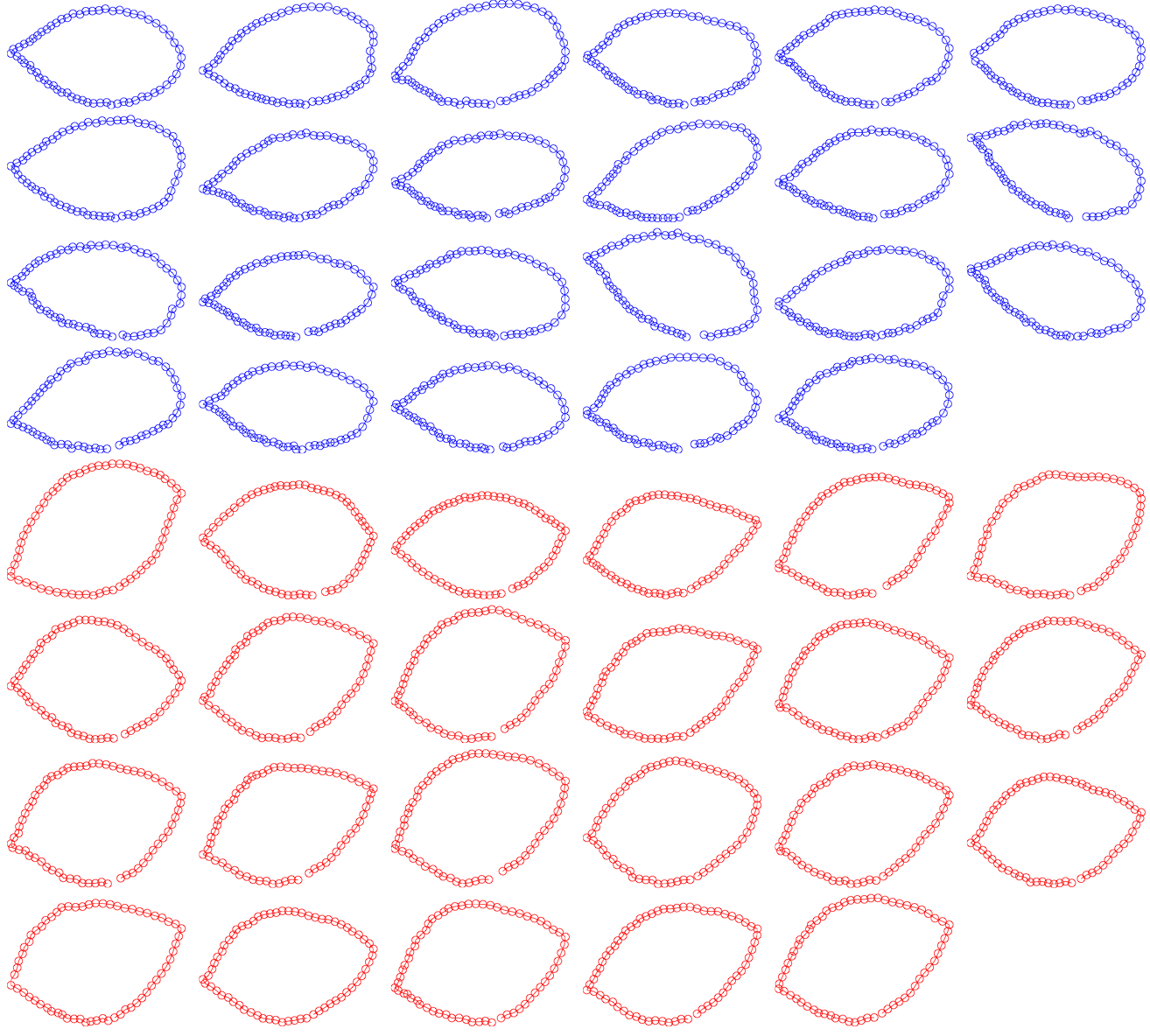}}
    \scalebox{0.175}{\centerline{\includegraphics{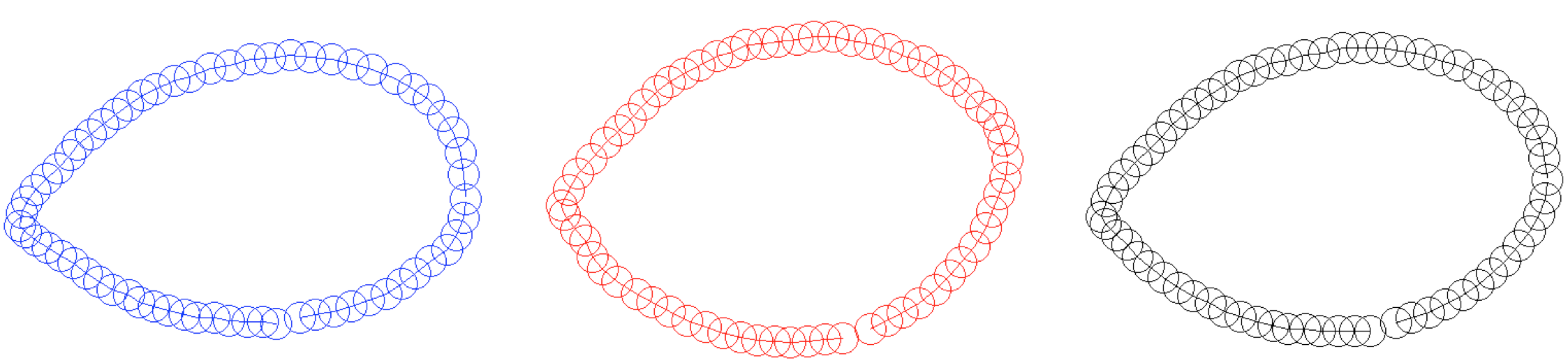}}}
    \vskip -0.1in
    \caption
    {
      {\bf Tree Leaf Database.}
      {\bf Top: } shows the 2 groups of leaves from 2 different species of trees; one in red and the other in blue.
      {\bf Bottom: } shows the expected values for the group-mean shape variables $M_1$ (blue dots) and $M_2$ (red dots) and the optimal population
      mean parameter $M$ (black dots), after EM inference.
    }
    \label {fig:leafMean}
  \end{center}
\end{figure}

This section shows results of the proposed hierarchical multigroup shape modeling and inference on a real dataset, namely the Tree Leaf
Database\cite{LEAF} comprising images of leaves of 90 wood species growing in the Czech Republic. We used 2 of the largest available groups
comprising the species (i)~\emph {Carpinus betulus} having 23 leaf images (Figure~\ref{fig:leafMean}; blue group) and (ii)~\emph {Fagus sylvatica}
also having 23 leaf images (Figure~\ref{fig:leafMean}; red group). The leaf stem was removed from the images manually. Interestingly, while the blue
group has oblong leaves that have high curvature at one end and low at the other, the red group has oblong leaves that are more symmetric with similar
curvatures at both ends.

{Preprocessing: } All leaf data pointsets $\{ x_i \}, \{ y_j \}$ initially undergo Procrustes alignment\cite{Goodall1991} to remove the effects of
translation, scale, and rotation.

\begin{figure}[!t]
  \begin{center}
    \centerline{\includegraphics[width=\columnwidth]{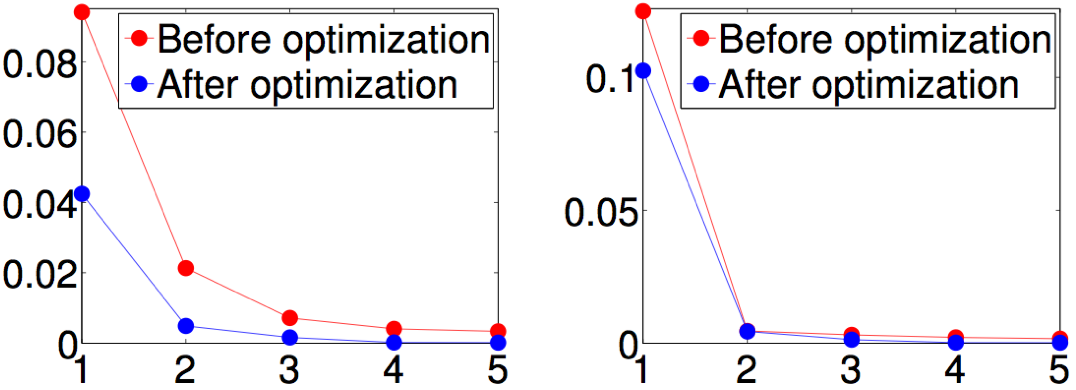}}
    \centerline{\includegraphics[width=\columnwidth]{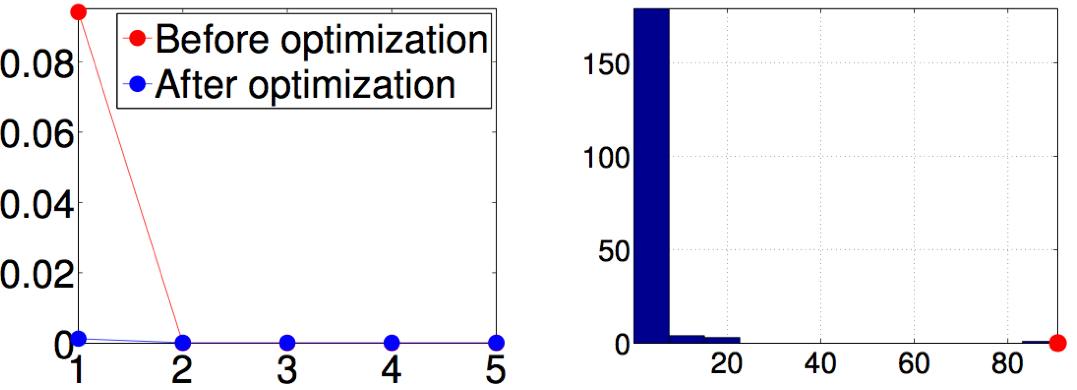}}
    \vskip -0.1in
    \caption
    {
      {\bf Left Column, (Top-to-Bottom): } show the largest $5$ eigen values of the covariance
      matrices $C_1$ and $C_2$, respectively, before and after EM optimization.
      {\bf Right Column, Top: } show the largest $5$ eigen values of the covariance matrices $C$, before and after EM optimization.
      {\bf Right Column, Bottom: } shows the histogram of Hotelling's $T^2$ statistics obtained by permutation testing (200 randomly-chosen group-label
      permutations) and the $T^2$ statistic for the non-permuted groups (red dot).
    }
    \label {fig:leafCovTest}		
  \end{center}
\end{figure}

% result: mean
Figure~\ref{fig:leafMean} also shows the population shape mean (black) and the expected values of the group shape means (blue, red). The expected
values of the shapes for the 2 groups clearly show the aforementioned subtle difference in curvatures in the leaves of the 2 species.
% result: covariance
Figure~\ref{fig:leafCovTest} shows the covariances for the leaf groups and population after EM optimization. We see that the group covariances
$C_1,C_2$ and the population covariance $C$ after EM optimization indicate lower variance.

\section {Application I --- Hypothesis Testing}
\label {sec:testing}

After the multi-group model is fit to the data, we can perform hypothesis testing to compare any pair of groups. Since the distribution of each group
is modeled by a Gaussian, e.g. $G (\cdot; M_1,C_1), G (\cdot; M_2,C_2)$, we use Hotelling's two-sample $T^2$ statistic to measure dissimilarity
between any pair of groups. Conventionally, the p value associated with the hypothesis test is obtained by exploiting the relationship of the $T^2$
statistic to the F distribution with a certain degrees of freedom depending on the sample sizes (e.g. $I, J$, in our case) and the dimensionality ($T
D$). In our case, however, the sample sizes ($I+J$) can be far lower than the dimensionality $TD$, which prevents the use of the F distribution for
computing p values. Thus, we propose \emph {permutation testing}, using the $T^2$ as the test statistic, for hypothesis testing to overcome the the
problem of low sample size.

\subsection {Validation on Simulated Data}

% result: permutation test
For the simulated box-bump dataset, Figure~\ref{fig:boxBumpCovTest} shows the expected result of permutation testing (H$_0$: the 2 group models are
the same) indicating the the non-permuted labeling produces the most extreme statistic (p value $= 1/35 = 0.0286$) thereby leaning significantly
towards rejection of the null hypothesis underlying permutation testing.

\subsection {Results on a Database of Leaves}

% result: permutation test
Figure~\ref{fig:leafMean} shows the 2 leaf groups that presents a challenge for hypothesis testing because the differences between the leaves are
subtle. Figure~\ref{fig:leafCovTest} shows the results of permutation testing indicating the the non-permuted labeling produces the most extreme
statistic (p value $= 0.005$) thereby leaning significantly towards rejection of the null hypothesis underlying permutation testing. This is the
desired result because the two leaf groups were known to correspond to two different tree species.

\section {Application II --- Classification}
\label {sec:classification}

After the multi-group model is fit to \emph{training} data, we can classify unseen shapes as follows.
The \emph {test} pointset is first Procrustes aligned to the population mean yielding a pointset, say, $z$. Let the hidden random variable
corresponding to the test shape be $w$. Then, we evaluate the probability of the unseen shape $z$ being drawn from the each group model, i.e., $P (z |
M_1,C_1,\beta_1)$ and $P (z | M_2,C_2,\beta_2)$ and classify $z$ to the class that yields a higher probability. We can evaluate the aforementioned
probabilities as follows:
\begin{align}
  &
  P (z | M_1,C_1,\beta_1)
  =
  \int_w
  P (z, w | M_1,C_1,\beta_1) dw
  \nonumber \\
  &
  =
  \int_w
  P (z | w, M_1,C_1,\beta_1)
  P (w |    M_1,C_1,\beta_1)
  dw
  \nonumber \\
  &
  \approx
  \sum_{s=1}^S
  \frac {1} {S}
  P (z | w^s)
  \text { where }
  w^s \sim P (W | M_1,C_1,\beta_1).
\end{align}

\subsection {Results on a Database of Leaves}

We performed the classification task by training using only 3 leaves from each group and testing on the remaining 20 leaves in each group. We obtained
a correct-classification rate of $97.5 \%$.

\section {Discussion and Conclusion}
\label {sec:conclusion}

% hierarchical model
This paper propose a novel hierarchical graphical generative statistical model for modeling shapes (a shape is defined as an equivalence class of
pointsets\cite{Kendall1984}) within a population for modeling multiple groups within a population.
% inference
Model inference proceeds using the EM framework where the individual shape variables as well as the group-mean shape variables are treated as hidden
random variables.
% sampling
In the E step, the approximation of the Q function using Monte-Carlo simulation entails sampling in Kendall shape space for which we propose a novel
method that incorporates the (i) efficient HMC sampling coupled with (ii)~gradient projection on shape manifolds.
% validation, applications
We validate the proposed modeling and inference scheme on simulated box-bump shapes\cite{Kotcheff1998} and exploit the framework in two different
applications, namely hypothesis testing and classification, where we obtain very promising results on a difficult real database. The framework could
be extended for unsupervised learning of classes of shapes.

% NOT a Gaussian graphical model
The proposed hierarchical graphical model is \emph {not} a simple Gaussian graphical model because of (i)~smoothness penalty on individual shape
variables as well as (ii)~the (nonlinear) current norm used to model the joint distribution of the observed pointset and the associated hidden shape
variable. In this way, the proposed framework solves a non-trivial inference problem.

% C_1,C_2 don't depend on population parameters
This paper does \emph {not} model dependencies of the per-group covariances $C_1,C_2$ on appropriate
population-level parameters. It also ignores smoothness parameters on the group means because of the
limited practical use of these parameters given many observed datasets per group.
These limitations can lead to areas for future work. Nevertheless, we believe that the paper still
makes important contributions.

% In the unusual situation where you want a paper to appear in the references without citing it in the main text, use \nocite{langley00}
\bibliographystyle{plain}
\bibliography{Bibtex_Shape}

\end{document}